\documentclass[a4paper,fleqn]{cas-sc}

\usepackage{picinpar}
\usepackage{cutwin}
\usepackage{graphicx}
\usepackage[utf8]{inputenc} 
\usepackage[T1]{fontenc}    
\usepackage{url}            
\usepackage{booktabs}
\usepackage{caption}
\usepackage{rotating}
\usepackage{makecell}
\usepackage{cleveref}
\usepackage{algpseudocode}  
\usepackage{amsmath,amssymb,amsfonts}
\usepackage{xcolor}
\usepackage{array}
\usepackage[ruled, linesnumbered]{algorithm2e}
\usepackage{bm}

\usepackage{float}  
\usepackage{wrapfig}
\usepackage{apacite}
\usepackage[numbers,square]{natbib}
\usepackage{hyperref}  
\usepackage{subfigure}

\newcolumntype{L}[1]{>{\raggedright\let\newline\\\arraybackslash\hspace{0pt}}m{#1}}
\newcolumntype{C}[1]{>{\centering\let\newline\\\arraybackslash\hspace{0pt}}m{#1}}
\newcolumntype{R}[1]{>{\raggedleft\let\newline\\\arraybackslash\hspace{0pt}}m{#1}}

\newcommand{\mname}{RIDA\xspace}

\SetKwInput{KwInput}{Input}                
\SetKwInput{KwOutput}{Output}
\SetKwFunction{Enum}{Enumerate}

\begin{document}
\let\WriteBookmarks\relax
\def\floatpagepagefraction{1}
\def\textpagefraction{.001}

\title{\mname: A Robust Attack Framework on Incomplete Graphs}    
\shorttitle{\mname: A Robust Attack Framework on Incomplete Graphs}
\tnotemark[1,2]
\author[1]{Jianke Yu}[orcid=0000-0002-2032-7727]
\ead{jianke.yu@student.uts.edu.au}
\shortauthors{Jianke Yu et~al.}
\credit{Conceptualization, Writing – original draft, Methodology, Software, Investigation}

\author[1]{Hanchen Wang}[orcid=0000-0003-3158-9586]
\ead{hanchen.wang@uts.edu.au}
\credit{Conceptualization, Writing - Review \& Editing, Methodology, Supervision}
\cortext[cor1]{Corresponding author}

\author[2]{Chen Chen}[orcid=0000-0003-3908-6545]
\ead{chenc@uow.edu.au}
\credit{Resources, Writing - Review \& Editing, Formal analysis}

\author[3]{Xiaoyang Wang}[orcid=0000-0003-3554-3219]
\ead{xiaoyang.wang1@unsw.edu.au}
\credit{Data Curation, Writing - Review \& Editing}

\author[1]{Lu Qin}[orcid=0000-0001-6068-5062]
\ead{lu.qin@uts.edu.au}
\credit{Conceptualization, Writing - Review \& Editing, Project administration}

\author[3]{Wenjie Zhang}[orcid=0000-0001-6572-2600]
\ead{wenjie.zhang@unsw.edu.au}
\credit{Conceptualization, Writing - Review \& Editing, Project administration}

\author[1]{Ying Zhang}[orcid=0000-0002-2674-1638]
\ead{ying.zhang@uts.edu.au}
\credit{Funding acquisition, Supervision}

\author[4]{Xijuan Liu}
\ead{liuxijuan@zjgsu.edu.cn}
\cormark[1]
\credit{Writing - Review \& Editing}

\affiliation[1]{organization={University of Technology Sydney},
    city={Ultimo},
    postcode={2007}, 
    state={NSW},
    country={Australia}
}
\affiliation[2]{organization={University of Wollongong},
    city={Wollongong},
    postcode={2522}, 
    state={NSW},
    country={Australia}
}
\affiliation[3]{organization={University of New South Wales},
    city={Kensington},
    postcode={2052}, 
    state={NSW},
    country={Australia}
}
\affiliation[4]{organization={Zhejiang Gongshang University},
    city={Hangzhou},
    postcode={310018}, 
    country={China}
}

\newcommand{\mypara}[1]{\vspace{1mm} \noindent \textbf{#1}}

\begin{abstract}
As Graph Neural Networks (GNNs) are increasingly vital, they are becoming primary targets for poisoning attacks. 
These attacks inject perturbations into the datasets used for retraining, resulting in a decline in model performance.
To create robust GNN models, it is crucial to design strong poisoning attack models as foundational benchmarks and guiding references.
Existing attack models are based on the assumption that attackers have access to the complete graph structure and attribute information.
However, in many scenarios, only partial or incomplete graph data are available due to the data source's privacy policies.
Given this, we propose a practical attack method for incomplete graphs, named Robust Incomplete Deep Attack Framework (\mname).
It is the first approach for robust gray-box poisoning attacks on incomplete graphs.
To ensure a reliable surrogate model for attacks, we incorporate a Depth-plus GNN module for long-range information propagation and a Local-global Aggregation module to refine feature aggregation.
Additionally, we optimize and execute poisoning attacks on incomplete graphs through the Holistic Adversarial Attack module.
Extensive evaluations against 9 state-of-the-art baselines on 3 real-world datasets demonstrate that \mname outperforms existing methods in attacking GNNs on incomplete graphs.
\end{abstract}


\maketitle

\section{Introduction}
\label{sec:intro}
Graph Neural Networks (GNNs) play an increasingly vital role in contemporary data science.
Their advanced learning capabilities significantly impact various domains, 
such as social network analysis~\citep{zhang2022robust,bao2024graph}, recommendation systems~\citep{chen2024macro,li2024graph,song2021cbml}, financial risk control~\citep{xu2022payment}, database enhancement~\citep{li2023graphix,sun2024breaking,yan2024gidcl}, trajectory data mining~\citep{chen2024deep} and anomaly and fraud detection ~\citep{yu2023group,nguyen2023example}.
As GNNs are being deployed in multiple security-sensitive areas, concerns regarding the security and robustness of GNNs against various attacks are growing~\citep{liu2023everything,wang2023scapin,hussain2021structack,zhao2021robust}.
Among these threats, poisoning attacks on attributed graphs pose significant challenges by subtly manipulating the structure of the training graph to degrade model performance. 
These attacks are especially difficult to defend since GNN models need to be regularly retrained to adapt to evolving graph structures and dynamic environments~\citep{liu2022towards, 10.1145/3589335.3651501}.
By injecting maliciously crafted perturbations into the data, these attacks can compromise the performance of GNN models, leading to errors in prediction and classification tasks~\citep{zou2021tdgia,liu2022gradients}.
In response, research efforts have notably increased in this area.
Numerous poisoning attack methods on attributed graphs have been proposed as foundational benchmarks and guiding references, helping in the development of secure and robust GNN models~\citep{zugner2018adversarial,zhang2023minimum,xiao2023black,10.1145/3589335.3651501}.

In this paper, we focus on gray-box poisoning attacks, which are one of the most significant poisoning attack strategies~\citep{liu2022towards,lin2020exploratory,xu2019topology,zügner2018adversarial,zugner2018adversarial}.
In gray-box scenarios, attackers have partial knowledge of the target model, including its training data and input-output behavior, but lack full access to its internal parameters and complete model architecture.
Current state-of-the-art attack methods use a loss function designed to maximize the training loss of target GNN models, directly opposing their regular training objectives~\citep{zugner2018adversarial,xu2019topology}. 
In these methods, attackers treat the adjacency matrix as a learnable parameter.
By applying backpropagation to compute gradients with respect to the adjacency matrix, attackers can identify a minimal set of optimal edges to perturb, disrupting the correlation between attributes and graph structure, and thereby degrading the performance of the target model.
However, current research is under the assumption that the attackers have access to the complete graph information, which is usually not practical in many real-world applications. 
For instance, in social networks, it is difficult for attackers to obtain complete data because some users tend to hide their private information, such as gender, age, health condition, home address, and financial status.
Therefore, solely defending against existing attack models is insufficient when designing robust models for real-world applications.

The incomplete attribute information has a great impact on the performance of current attack methods, which has been validated by our extensive experiments (more details can be found in~\Cref{fig:nrates} and~\Cref{sec:furtheranalyses}). 
Through these experiments, we observed that as the proportion of missing vertex attributes increases, the performance of existing state-of-the-art attack methods decreases substantially. 
The declines occur because existing methods are sensitive to the overall distribution of attributes and inherently require attacking the target model on a global scale.
More specifically, the attack process in current state-of-the-art models relies on a well-fitted surrogate model built from the available graph information, which simulates the behavior of the target model for the downstream attack module.
However, when certain attributes are missing, the reliability of the surrogate model is compromised.
The surrogate model would easily introduce noise into the poisoning attack processing because of the incompleteness of the graph, which disrupts the correlation between attributes and structure.
As a result, the effectiveness of attack models will decline sharply.
Therefore, effective attack methods specifically designed to handle such scenarios are needed to assist in the development of more robust GNN models.
To fill the gap, in this paper, we introduce a new model called  \underline{R}obust \underline{I}ncomplete \underline{D}eep \underline{A}ttack Framework (\mname), which enables effective poisoning attacks on graphs with incomplete attribute information.
\mname addresses two key challenges in gray-box poisoning attacks on incomplete graphs: the failure of the surrogate model and the instability of the attack process.

\begin{itemize}
    \item \textbf{The failure of the surrogate model.} 
    The surrogate model simulates the behavior of the target model, enabling attackers to approximate its predictions or decision-making process without direct access. 
    In ideal scenarios where full graph information is available, the surrogate models within existing attack methods can closely mimic the training and inference processes of the target model. 
    However, conventional surrogate models are not designed for incomplete graphs.
    Missing data disrupts the information propagation process, causing these models to perform poorly in such scenarios.
    \item \textbf{The instability of the attack process.} The instability of the attack process arises from conventional attack methods' heavy reliance on the correlation between vertex attributes and graph structure.
    The target model achieves high accuracy by leveraging these correlations, and poisoning attacks typically aim to identify and disrupt such correlations. 
    However, these correlations are weakened or inaccurate in incomplete graphs, rendering conventional attack strategies ineffective in such scenarios.
\end{itemize}


To overcome these challenges, \mname incorporates three modules: the Depth-plus GNN module, the Local-global Aggregation module, and the Holistic Adversarial Attack module. 
The Depth-plus GNN module enhances the surrogate model by propagating information across the entire graph, leveraging the available information to compensate for missing data.
The Local-global Aggregation module refines the model's attention mechanism during propagation by considering information from both global and local perspectives, thus improving the reliability of the surrogate model. 
This module captures how features change between the current and previous layers from the local perspective and enables the propagation process to retain the original vertex attribute information from the global perspective.
It helps the model perform stable long-range propagation and optimizes the surrogate model used in attacks.
Finally, the Holistic Adversarial Attack module builds on the previous modules and focuses on optimizing the vertex features for the attack process, enhancing poisoning attack effectiveness on incomplete graphs.
The main contributions of this work are summarized as follows: 
\begin{itemize}
    \item To the best of our knowledge, \mname is the first model to achieve effective poisoning attacks on incomplete graphs. 
    The model demonstrates robust attack performance in real-world scenarios where attributes of a significant proportion of vertices are missing.
    By releasing \mname, we provide GNN designers with a more realistic attack scenario to improve the robustness of GNN models.
    \item \mname introduces three well-designed key modules to address existing challenges. 
    The Depth-plus GNN module improves long-range information propagation, addressing the failure of surrogate models on incomplete graphs. 
    The Local-global Aggregation module refines feature aggregation to further optimize the performance of surrogate model. 
    The Holistic Adversarial Attack module optimizes and applies perturbations, enabling the final poisoning attacks on incomplete graphs. 
    \item 
    To evaluate the performance of the proposed model, comprehensive experiments are conducted 
    against nine state-of-the-art baselines on three real-world datasets. 
    The experimental results demonstrate that \mname significantly outperforms existing methods.  
    Extensive experiments, including ablation studies and additional analyses, further validate the effectiveness of each component of \mname.
\end{itemize}

\section{Preliminary}
This section covers the essential concepts for \mname, including definitions of attributed and attribute-incomplete graphs, an overview of Graph Neural Networks (GNNs) as target models, and the problem definition for gray-box poisoning attacks on incomplete graphs.


\mypara{Attributed graph.}
An attributed graph is denoted as \(\mathcal{G} = (V, E, \bm{X})\), where \(V = \{v_1, v_2, \dots, v_n\}\) represents the set of vertices, \(E\) denotes the set of edges, and \(\bm{X}\) is the attribute matrix, with each row corresponding to the attributes of a vertex in \(V\).
Additionally, we define $\bm{A} \in \{0, 1\}^{|V| \times |V|}$ as the adjacency matrix of the graph $\mathcal{G}$, where $|V|$ denotes the number of vertices in the graph.

\mypara{Attribute-incomplete graph.}
An attribute-incomplete graph is denoted as \( \mathcal{G}' = (V, E, \bm{X}^{\phi}) \).
In this graph, \( V \) and \( E \) remains unchanged, while \( \bm{X}^{\phi} \) is the attribute matrix with certain attributes missing.
The incompleteness of \( \bm{X}^{\phi} \) is characterized by two parameters: \( \alpha \), the proportion of missing attributes per vertex, and \( \beta \), the proportion of vertices with incomplete attributes. 

\mypara{Graph Neural Networks.}
The attack targets Graph Neural Networks (GNNs).
Numerous outstanding GNN models are widely used across various domains~\citep{wu2019simplifying,wang2022neural,yu2024temporal}. 
The structure of these models can be generally summarized as follows:
\begin{equation}
\label{eq:gnn}
    {\bm h}^{(k)}_u = \mathcal{COM}^{(k)}({\bm h}^{(k-1)}_u, \mathcal{AGG}^{(k)}\{{\bm h}^{(k-1)}_{u'}; u' \in N(u)\}),
\end{equation}
where ${\bm h}^{(k)}_u$ denotes the $k$-th layer's representation of vertex $u$. 
$\mathcal{AGG}$ is the function that iteratively updates the representation of a vertex by aggregating the features of its neighboring vertices.
$\mathcal{COM}$ is the combination operation that updates the representation of vertex $u$ by merging the aggregated neighbor representations with its previous layer representation ${\bm h}^{(k-1)}_u$.
The differentiator between GNNs is their unique aggregation and combination methods.



\mypara{Problem definition.} We focus on gray-box poisoning attacks on attribute-incomplete graphs.
Following previous works~\citep{10.1145/3589335.3651501,xu2019topology,zügner2018adversarial}, these attacks are carried out by perturbing a limited subset of edges.
The perturbed graph, \( \mathcal{G}_p = (V, E_p, \bm{X}) \), misleads the target model into developing a lower-quality model, where \( E_p \) denotes the set of edges after perturbation.
The proportion of altered edges between \( E_p \) and \( E \) is less than \( \epsilon \), ensuring the stealthiness of the attack. 
The attack operates in a gray-box setting, where the attacker only has access to 
an attribute-incomplete graph \( \mathcal{G}' \), and the set of vertex labels \( \bm{Y} \), with \( |\bm{Y}| \ll |V| \).
\mname is specifically designed to degrade the performance of GNN models globally.
In other words, instead of targeting specific vertices, our approach focuses on reducing the overall performance of the model.
\begin{figure*}[tb]
    \centering
    \includegraphics[width=\linewidth]{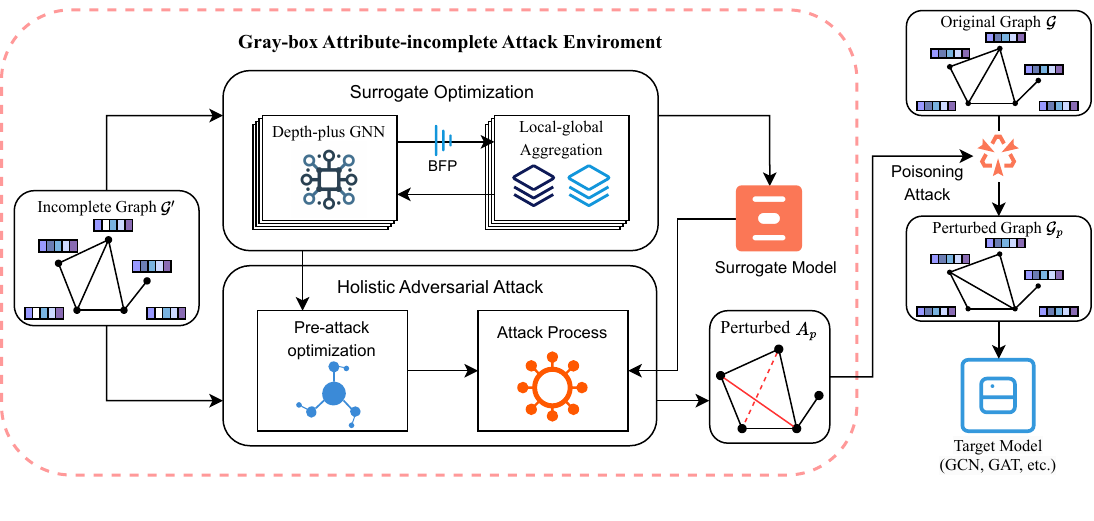}
    \caption{The framework of \mname}
    \label{fig:framework}
\end{figure*}
\section{Model}

This section provides a detailed explanation of \mname. 
\Cref{fig:framework} provides an overview of the model.
Specifically, \mname first integrates the Depth-plus GNN and Local-global Aggregation modules to enhance the surrogate model for incomplete graphs, then applies the Holistic Adversarial Attack module to optimize and perform poisoning attacks in this incomplete-graph scenario.
Finally, the target model is trained on the structure-perturbed graph, leading to a degradation in its performance.


\subsection{Depth-plus GNN Module}

The purpose of this module is to address the failure of conventional surrogate models on incomplete graphs.
To achieve this, it maximizes the utilization of available information.
Specifically, we leverage GNN techniques to propose an effective long-range information propagation strategy for incomplete graphs.

Currently, many high-performing spectral graph neural networks have been developed. 
These models can effectively propagate and integrate information between vertices. 
The Graph Convolutional Network (GCN), as a representative of these models, aggregates each vertex's information with its neighboring vertices layer by layer through multiple convolutions, thereby propagating information across the entire graph~\citep{kipf2017semisupervised}. 
The propagation rule for GCN is implemented as follows:
\begin{equation}
\bm{H}^{(l+1)} = \sigma(\bar{\bm{D}}^{-\frac{1}{2}}\bar{\bm{A}}\bar{\bm{D}}^{-\frac{1}{2}}\bm{H}^{(l)}\bm{W}^{(l)}),
\end{equation}
where $\bm{H}^{(l)}$ is the $l$-th layer feature matrix, $\bar{\bm{A}}$ is the adjacency matrix with added self-loops, $\bar{\bm{D}}$ is the degree matrix of $\bar{\bm{A}}$, $\bm{W}^{(l)}$ is the learnable weight matrix of $l$-th layer and $\sigma$ is nonlinear activation function. 
Traditional GNNs struggle to aggregate attributes from distant vertices, as information tends to become indistinguishable across multiple layers.
To mitigate the impact of this issue, we introduce the Depth-plus GNN in our model.
Initially, the module initializes the vertex features as follows:
\begin{equation}
  \bm{X}'_{ij} = \begin{cases} 
    0 & \text{if } \bm{X}^{\phi}_{ij} \text{ is missing} \\ \bar{\bm{X}}^{\phi}_{ij} \cdot \omega + (1-\omega) & \text{otherwise},
  \end{cases}
\end{equation}
where $\bar{\bm{X}}^{\phi}$ is the normalized $\bm{X}^{\phi}$, $\bm{X}^{\phi}_{ij}$ and $\bar{\bm{X}}^{\phi}_{ij}$ denotes the element in the $i$-th row and $j$-th column of matrix $\bm{X}^{\phi}$ and $\bar{\bm{X}}^{\phi}$, respectively, and $\omega$ represents a weighting factor that distinguishes between missing and existing attributes.
This strategy safeguards the original attributes from noise during propagation.
Then, we simplify the information propagation process into the following form:
\begin{equation} \label{eq:sgc}
\bm{H}^{(l+2)} = \tilde{\bm{A}}(\tilde{\bm{A}}\bm{H}^{(l)}\bm{W}^{(l)})\bm{W}^{(l+1)} = \tilde{\bm{A}}^2\bm{H}^{(l)}\bm{W},
\end{equation}
where $\tilde{\bm{A}} = \bar{\bm{D}}^{-\frac{1}{2}}\bar{\bm{A}}\bar{\bm{D}}^{-\frac{1}{2}}$ and $\bm{W}$ is a weight matrix, which can be seen as the result of multiplying the weight matrices from all layers.
After removing the nonlinear activation function, the model can retain as much of the original information as possible during the propagation process while improving propagation efficiency.
To further enhance the module's ability to aggregate long-range information, the process in~\Cref{eq:sgc} is divided into feature propagation and feature transformation parts.

\mypara{Feature propagation.} 
This part utilizes the adjacency matrix for a controlled long-range information propagation process. 
The specific implementation is as follows:
\begin{equation} \label{eq:ak}
\begin{aligned}
\bm{X}^{(k)} &= 
\begin{cases}
\bm{X}' & \text{if } k = 0 \\
\lambda\hat{\bm{A}}\bm{X}^{(k-1)} + (1-\lambda)\bm{X}^{(k-1)} & \text{if } k > 0,
\end{cases}
\end{aligned}
\end{equation}
where \(\hat{\bm{A}}\) represents the normalized adjacency matrix excluding self-loops. 
The parameter \(k \in \mathbb{Z}_{\geq 0}\) is the number of aggregation layers. 
The decay parameter \(\lambda= \delta (1-\gamma)^k\) is determined by three factors: the hyperparameters \(\delta\) and \(\gamma\), along with the layer count \(k\). 
To avoid overemphasizing the features of the target vertex during the information propagation process, the target vertex's features are handled separately instead of preprocessing the adjacency matrix with a self-loop.
On the other hand, thanks to the decay parameter, the module can control the influence of features from neighboring vertices at varying distances, adjusting how much each neighbor’s features contribute to the final aggregation based on their relative distance. 
This flexibility allows the model to prioritize more relevant features while minimizing the impact of distant or less relevant information. 
As a result, this strategy ensures effective and efficient long-range propagation on incomplete graphs.

\mypara{Feature transformation.}
The feature transformation part maps the result of propagation \(\bm{X}^{(K)}\) to achieve prediction. 
This process is implemented using a series of multi-layer linear transformations:
\begin{equation}
\hat{\bm{Y}} = \bm{W}_1(\bm{W}_2 \bm{X}^{(K)} + \bm{b}_1) + \bm{b}_2,
\end{equation}
where \(\bm{W}_1, \bm{W}_2, \bm{b}_1, \bm{b}_2\) are learnable weight matrices and biases, respectively.
The prediction results $\hat{\bm{Y}}$ is utilized to train the surrogate model in the attack process.
Since the surrogate model is only used to guide the attack process, nonlinear activation and normalization are not required for the results.
Following the above forward propagation and transformation, the parameters are optimized by minimizing the cross-entropy loss function.

By the above strategy, the module achieves long-range information propagation while maintaining the model's efficiency.
In particular, Depth-plus GNN maintains a time complexity of $O(|E|)$ for attribute propagation per layer, where $|E|$ denotes the number of edges.
Furthermore, since both the vertex attributes and the adjacency matrix remain unchanged during this phase, the propagation step only needs to be computed once prior to the training process.
This eliminates the need for repeated calculations and optimizes the efficiency of the module.

\subsection{Local-global Aggregation Module}

The Local-global Aggregation module is also designed to address the failure of the surrogate model on incomplete graphs. 
Specifically, it enhances the Depth-plus GNN module by refining the aggregation process through local and global perspectives.
The local perspective captures how features evolve between consecutive layers, ensuring that immediate, short-range information is effectively propagated. 
Meanwhile, the global perspective helps retain the original vertex attribute information, allowing the model to maintain consistency over longer propagation distances. 
This module first employs a bifocal feature processor, which is designed to prioritize and retain the original attribute information during the aggregation process.
Then, using an attention mechanism, it performs aggregation from both local and global perspectives simultaneously.

\mypara{Bifocal Feature Processor (BFP).}
In each round of propagation, missing features at a vertex are supplemented by those propagated from other vertices.
Since these supplemented features do not contain the vertex's own initial information, treating them as regular features in subsequent attention calculations could risk destabilizing the model's aggregation process. 
As a result, the Bifocal Feature Processor is introduced to selectively ignore these supplemented features when calculating attention coefficients. 
Initially, the vertices are divided into two categories: attribute-complete and attribute-incomplete. 
Then, the vertices from each category participate independently in aggregation. 
Additionally, for the attribute-incomplete category, the supplemented features are masked out during the attention coefficient calculation to ensure stable aggregation.

\mypara{Local-global aggregation.}
To optimize the performance of the feature propagation component of the Depth-plus GNN module, this module employs an attention mechanism to further enhance feature aggregation from both local and global perspectives.
Specifically, the module first applies the bifocal feature processor to the propagation results of each layer \(\bm{X}^{(k)}\) (where \(\bm{X}^{(k)}\) is defined in~\Cref{eq:ak}), producing \(\bm{X}_b^{(k)}\) and \(\bm{X}_{b'}^{(k)}\). 
Here, \(\bm{X}_b^{(k)}\) and \(\bm{X}_{b'}^{(k)}\) represent the \(k\)-th layer feature matrices corresponding to the attribute-complete and attribute-incomplete category vertices, respectively.
Similarly, \(\bm{X}'\) is also divided into \(\bm{X}'_b\) and \(\bm{X}'_{b'}\).
Note that the aggregation process for the two categories is identical and independent.
For clarity, we will only describe the operations on $\bm{X}_b^{(k)}$.
After the above process, the module calculates the cosine distance between \(\bm{X}^{(k)}_b\) and \(\bm{X}^{(k-1)}_b\) for \(k > 0\), and between \(\bm{X}^{(k)}_b\) and \(\bm{X}^{(0)}_b\) for each layer, to obtain the local-global attention coefficients:
\begin{equation}
    \bm{C} = \frac{{\bm{X}^{(k)}_b \cdot \bm{X}^{(k-1)}_b}}{{\|\bm{X}^{(k)}_b\| \cdot \| \bm{X}^{(k-1)}_b\|}} \cdot \frac{{\bm{X}^{(k)}_b \cdot \bm{X}^{(0)}_b}}{{\|\bm{X}^{(k)}_b\| \cdot \| \bm{X}^{(0)}_b\|}},
\end{equation}
where $\|\cdot\|$ denotes L2 norm computation. 
The attention coefficients obtained in this strategy can further optimize the feature propagation process. 
To be precise, local attention guides the model in preserving important information during long-range propagation, while global attention helps the model retain the original vertex attribute information.
Finally, the feature propagation process can be optimized using the attention coefficients, and the features in each layer are aggregated as follows:
\begin{equation} \label{eq:nak}
\begin{aligned}
\bm{X}_b^{(k)} &= 
\begin{cases}
\bm{X}'_b & \text{if } k = 0 \\
\lambda\bm{C} \hat{\bm{A}}\bm{X}^{(k-1)}_b + (1-\lambda\bm{C})\bm{X}^{(k-1)}_b & \text{if } k > 0.
\end{cases}
\end{aligned}
\end{equation}
The feature matrix of attribute-incomplete category vertices \(\bm{X}^{(k)}_{b'}\) can be obtained in the same way.
By merging \(\bm{X}^{(k)}_b\) with \(\bm{X}^{(k)}_{b'}\), the final feature propagation result for this layer \(\bm{X}^{(k)}\) can be achieved.
This aggregation strategy allows the model to effectively manage attribute information across varying distances. 

\subsection{Holistic Adversarial Attack Module}
This module is designed to perform attacks on incomplete graphs.
As discussed in~\Cref{sec:intro}, the correlations between vertex attributes and graph structure  can be unreliable in such scenarios. 
To mitigate this issue, the module optimizes features and then conducts adversarial attacks to implement optimal perturbations.
A poisoning attack is then performed to generate the perturbed graph \(\mathcal{G}_p\).
The objective of this process is defined by the following equation:
\begin{equation}
\label{eq:atktarget}
    \min_{\mathcal{G}_p \in \Phi(\mathcal{G})} \mathcal{L}_{\text{atk}}(f_{\theta^*}(\mathcal{G}_p)) \ \text{s.t.} \ \theta^* = \arg\min_{\theta} \mathcal{L}_{\text{train}}(f_{\theta}(\mathcal{G}_p)),
\end{equation}
where $f_{\theta}$ is the surrogate model function with parameters $\theta$, $\mathcal{L}_{\text{train}}$ is the loss function of this surrogate model, $\mathcal{L}_{\text{atk}} = -\mathcal{L}_{\text{train}}$ is the loss function the attacker seeks to minimize, and \( \Phi(\mathcal{G}) \) represents the set of allowed edge modifications in the graph \( \mathcal{G} \), resulting in the final perturbed graph \( \mathcal{G}_p \).

\mypara{Pre-attack optimization.}
The modules introduced above cannot be used directly as the surrogate model, as the process involves numerous power operations on the adjacency matrix during forward propagation, which makes backpropagation computationally prohibitive.
To address this challenge, we propose the Holistic Adversarial Attack module to enable an efficient attack process on incomplete graphs.
This module leverages the long-range feature propagation traces of the Depth-plus GNN module to optimize the features involved in the attack process, allowing the model to fully utilize the entire graph's information when attacking incomplete graphs.
More specifically, the module first constructs a set of intermediate adjacency matrices to preserve vertex connections:
\begin{equation}
    \begin{aligned}
    \bm{T}^{(k)} &= 
    \begin{cases}
    \hat{\bm{A}} & \text{if } k = 0 \\
    \hat{\bm{A}} \cdot \bm{T}^{(k-1)} & \text{if } k > 0.
    \end{cases}
    \end{aligned}
\end{equation}
Then, the module distinguishes and records the feature propagation weights at different distances:
\begin{equation}
    \bm{A}_{\phi}^{(k)} = \lambda \bm{T}^{(k-1)} + (1 - \lambda) \bm{A}_{\phi}^{(k-1)},
\end{equation}
where the decay parameter \(\lambda\) is consistent with the one used in the Depth-plus GNN module, \(\bm{A}_{\phi}^{(k)}\) is the propagation matrix at the $k$-th layer, and \(\bm{A}_{\phi}^{(0)} \) is an identity matrix.
Based on this propagation mechanism, the attack module optimizes the features involved in the attack process as follows:
\begin{equation}
\label{eq:xs}
    \bm{X}_s = \eta \bm{A}_{\phi}^{(K-1)} \bm{X}^{\phi}_n + (1 - \eta) \bm{X}^{\phi}_n,\\
\end{equation}
where \(\bm{X}_s\) represents the optimized features for the attack, \(\eta\) denotes the trade-off parameter, and \(\bm{X}^{\phi}_n\) is the normalized version of the original attributes \(\bm{X}^{\phi}\).
\(\bm{X}_s\) is the vertex attribute matrix with positional encoding. 
It incorporates the entire graph's structural information and enhances the module's effectiveness in attacking incomplete graphs.

\mypara{Attack process.} Finally, a GCN without nonlinear activation functions is used as the surrogate model during the attack stage. 
The poisoning attack is implemented by optimizing the adjacency matrix during training, as described in~\Cref{eq:atktarget}.
In detail, the module trains the surrogate model on the graph and minimizes the attacker's loss \( \mathcal{L}_{\text{atk}} \) to obtain the best-perturbed graph $\mathcal{G}_p$.
We reformulate this process as finding the optimal perturbation matrix \( \bm{A}_p \) to construct the adjacency matrix of \( \mathcal{G}_p \), where \( \bm{A}_{atk} = \bm{A} + \bm{A}_p \).
The perturbation matrix maximizes the attack performance while ensuring that the proportion of differing edges between the original structure \( \bm{A} \) and the perturbed structure \( \bm{A}_p \) remains below \( \epsilon \).

Initially, the module set up the perturbation matrix $\bm{A}_p^{(0)} = \mathbf{0}$. 
This all-zero adjacency matrix represents that no attacks have been conducted.
In the \(i\)-th iteration of the attack, we start by removing the diagonal elements from the perturbation matrix \(\bm{A}_p^{(i)}\) and ensuring that its values remain within the range from \(-1\) to \(1\).
The sign of the value is used to determine edge additions or deletions.
Subsequently, we construct the targeted adjacency matrix for the $i$-th attack round as $\bm{A}_m^{(i)} = \bm{A} + \bm{A}_p^{(i)}$. 
Afterward, the surrogate model is trained to obtain the optimized model parameters:
\begin{equation}  \label{eq:target}
    \theta^{(i)} = \arg\min_{\theta} \mathcal{L}_{\text{train}}(f_{\theta}(V, \bm{A}_m^{(i)}, \bm{X}_s, \bm{Y})).
\end{equation}
The module then uses $\theta^{(i)}$ to compute the gradient of $\bm{A}_m^{(i)}$ with respect to the attack loss function.
It represents the optimization direction of the perturbation matrix:
\begin{equation} \label{eq:attack}
    \nabla \bm{A}_m^{(i)} = \nabla_{\bm{A}_m^{(i)}} \mathcal{L}_{\text{atk}}(f_{\theta^{(i)}}(V, \bm{A}_m^{(i)}, \bm{X}_s, \hat{\bm{Y}})).
\end{equation}
After the computation, $\bm{A}_m^{(i)}$ is treated as the step size and multiplied by the gradient direction for matrix updates. 
Then, the diagonal elements of the matrix are set to 0, and all other elements are adjusted by subtracting the minimum value in the matrix to ensure non-negativity and avoid computational issues.
Finally, the position with the most significant change is chosen for this perturbation round. 
The \(i\)-th attack iteration is completed after storing the outcome in \(\bm{A}_P^{(i)}\).
To maintain graph connectivity, the module avoids attacking edges linked to vertices with a degree of \(1\). 
Note that we use the vertex labels \( \bm{Y} \) to train the surrogate model to simulate the target model, and the prediction results \( \hat{\bm{Y}} \) are used to perform the perturbation, reducing the impact caused by the mismatch between real labels and incomplete attributes.
After repeating the attack $|E| \times \epsilon$ times, the adjacency matrix of $\mathcal{G}_p$ can be obtained as:
\begin{equation}
    \bm{A}_{atk} = \bm{A} + \bm{A}^{(|E| \times \epsilon)}_p,
\end{equation}
where $\bm{A}^{(|E| \times \epsilon)}_p$ is the perturbation matrix, generated through $|E| \times \epsilon$ iterations of the edge perturbation process.

\section{Experiment}
\label{sec:exp}
In this section, we evaluate the effectiveness of \mname for gray-box poisoning attacks on incomplete graphs. 
We first describe the experimental setup, including datasets and target models, and then compare the model's attack performance with state-of-the-art methods under varying levels of missing information. 
Additionally, we conduct ablation studies to assess the impact of \mname's components and provide further analysis of its robustness and the effectiveness of its feature propagation processes.

\subsection{Experimenal Setup}

\mypara{Datasets.} 
We utilize three real-world datasets with attributes in those works: CORA~\citep{mccallum2000automating}, CORA-ML~\citep{mccallum2000automating} and CITESEER~\citep{sen2008collective}.
The CORA dataset, a widely used benchmark for graph-based learning, represents scientific papers as nodes and citation links as edges. 
It is commonly applied in graph-based semi-supervised learning and classification tasks. 
The CORA-ML dataset, derived from the original CORA dataset, serves as a citation network for evaluating machine learning models. Nodes correspond to scientific publications, are connected by citation relationships.  
The CITESEER dataset, another citation network, models scientific publications as nodes and citation links as edges.
The labels of nodes reflect different research domains.
To ensure fairness and comparability, we align our dataset setup with previous studies~\citep{liu2022towards,zügner2018adversarial}.
Only the largest connected components are utilized, and these datasets exhibit varied characteristics:
CORA consists of $2,485$ vertices, $5,069$ edges, $1,433$ dimensions for vertex attributes, and $7$ classes.
CORA-ML has $2,810$ vertices, $7,981$ edges, $2,879$ dimensions for vertex attributes and $7$ distinct classes.
CITESEER includes $2,110$ vertices, $3,757$ edges, $3,703$ dimensions for vertex attributes, and $6$ distinct classes.
They are split into labeled ($10\%$) and unlabeled ($90\%$) vertices.
The labels of the unlabeled vertices were kept hidden from both the attacker and the classifiers, being used exclusively to evaluate the performance of the models.
The code and datasets are available at \url{https://anonymous.4open.science/r/RIDA-EFC2}.

\mypara{Experimental environment.} 
In the experiments, we set the proportion of perturbed edges to \( \epsilon = 5\% \), the decay parameter hyperparameters to \( \gamma = 0.01 \) and \( \delta = 0.1 \), with \( \delta = 0.2 \) for CORA-ML. 
The trade-off parameter in~\Cref{eq:xs} is set to \( \eta = 0.05 \). 
The parameters \( \alpha \in \{0.1, 0.3\} \) and \( \beta \in \{0.7, 0.9\} \) control the incompleteness of vertex attributes, and the parameter distinguishing between missing and existing attributes is set to \( \omega = 0.9 \). 
The number of layers is set to \( K = 8 \) for CITESEER and \( K = 16 \) for other datasets.
The model's linear transformations include 2 layers, with a hidden layer dimension of 16 and a learning rate of 0.01.
The number of epochs for Depth-plus GNN and Holistic Adversarial Attack is $200$ and $100$, respectively. 
$\mathcal{L}_{\text{train}}$ employs cross-entropy loss. 
All experiments share a fixed seed and utilize the same data split configuration.
As for the target models, we chose GCN~\citep{kipf2017semisupervised}, GAT~\citep{veličković2018graph} and GraphSAGE~\citep{hamilton2017inductive} (referred to as SAGE in the result tables).
All target models share the same settings: 2 layers, a learning rate of 0.005, 200 epochs, and a hidden layer dimension of 16.
Following prior studies, the accuracy metric was chosen for evaluation. 
It represents the percentage of correct predictions out of the total number of predictions.
Each attack was executed ten times under each experimental setting, and the final result was obtained by averaging the performance after excluding the best and worst outcomes.
The experiments were conducted on a server running RHEL 8.8, equipped with an Intel Xeon E-2288G CPU, an NVIDIA RTX 6000 GPU with 24GB of memory, and 64GB of RAM.

\mypara{Compared methods.} We compared \mname with $9$ SOTA gray-box poisoning attack methods on attributed graphs, including:
DICE~\citep{waniek2018hiding}, EpoAtk~\citep{lin2020exploratory} (gray-box attack version), GraD~\citep{liu2022towards}, PGD~\citep{xu2019topology}, Meta-Self, Meta-Train, A-Meta-Self, A-Meta-Train, A-Meta-Both~\citep{zügner2018adversarial}.
We used their open-source code, keeping all parameters at their default settings. 
GraD and Meta-based methods utilized normalized vertex attributes to prevent gradient explosions.
We also tested the attack effectiveness of the recently published method DGA~\citep{10.1145/3589335.3651501}, but the target model's performance did not significantly change after being attacked by this method. 
Therefore, we do not report its results in our experiments.
To further validate the effectiveness of \mname, we compared it with three baselines based on the SOTA model A-Meta-Self, each using different data imputation strategies. 
The first method, ``MEAN,'' fills in missing features by using the average of all vertex features. 
The other two methods, ``KNN-10'' and ``KNN-100,'' impute missing features by averaging the features of the 10 and 100 nearest vertices based on cosine similarity, respectively.
Additionally, we tested a variant of \mname, referred to as \mname w/o FE.
This variant does not use \(\bm{X}_s\) during the attack phase but instead utilizes the original \(\bm{X}^{\phi}_n\).
It is used to evaluate the effectiveness of the optimization strategy in the Holistic Adversarial Attack module.

\begin{table*}[!ht]
    \centering
    \caption{Attack Efficacy of Various Algorithms at \(\alpha=0.1, \beta=0.7\)}
    \label{tab:n7f1}
    \resizebox{0.85\linewidth}{!}{\begin{tabular}{l|ccc|ccc|ccc}
    \toprule
    Dataset  & \multicolumn{3}{c|}{CITESEER} & \multicolumn{3}{c|}{CORA} & \multicolumn{3}{c}{CORA-ML} \\ \midrule
    Target Model & \multicolumn{1}{c}{GCN} & \multicolumn{1}{c}{GAT} & \multicolumn{1}{c|}{SAGE} & \multicolumn{1}{c}{GCN} & \multicolumn{1}{c}{GAT} & \multicolumn{1}{c|}{SAGE} & \multicolumn{1}{c}{GCN} & \multicolumn{1}{c}{GAT} & \multicolumn{1}{c}{SAGE} \\ \midrule
    Clean & 70.76\% & 73.33\% & 69.99\% & 83.39\% & 84.06\% & 82.74\% & 86.90\% & 86.54\% & 86.13\% \\ \midrule
    DICE & 69.98\% & 71.68\% & 68.84\% & 82.58\% & 82.65\% & 81.80\% & 85.80\% & 84.73\% & 85.26\% \\
    EpoAtk & 70.31\% & 72.34\% & 68.34\% & 82.15\% & 83.77\% & 81.61\% & 86.13\% & 85.95\% & 85.81\% \\
    GraD & 68.13\% & 70.94\% & 67.67\% & 81.09\% & 82.53\% & 81.94\% & 83.45\% & 83.41\% & 83.87\% \\
    PGD & 70.27\% & 71.62\% & 68.96\% & 81.51\% & 82.46\% & 81.61\% & 85.48\% & 85.00\% & 84.98\% \\
    Meta-Self & 68.59\% & 71.19\% & 68.36\% & 80.64\% & 82.21\% & 80.73\% & 83.08\% & 83.38\% & 84.09\% \\
    Meta-Train & 69.22\% & 71.59\% & 68.79\% & 81.48\% & 82.54\% & 81.32\% & 84.55\% & 84.84\% & 84.73\% \\
    A-Meta-Train & 70.94\% & 72.59\% & 69.19\% & 82.46\% & 83.75\% & 81.18\% & 84.94\% & 84.87\% & 84.32\% \\
    A-Meta-Self & 69.53\% & 71.25\% & 67.97\% & 80.55\% & 81.09\% & 80.52\% & 83.99\% & 83.18\% & 84.11\% \\
    A-Meta-Both & 71.72\% & 72.53\% & 69.34\% & 82.17\% & 83.42\% & 81.44\% & 85.43\% & 85.10\% & 84.87\% \\ \midrule
    KNN-10 & 68.65\% & 70.73\% & 68.10\% & 80.66\% & 81.73\% & 80.85\% & 84.03\% & 83.72\% & 84.38\% \\
    KNN-100 & 67.88\% & 71.18\% & 67.85\% & 81.00\% & 81.34\% & 81.01\% & 84.22\% & 83.24\% & 83.83\% \\
    MEAN & 69.44\% & 72.57\% & 68.65\% & 80.73\% & 80.89\% & 80.65\% & 84.02\% & 83.54\% & 84.36\% \\ \midrule
    \mname w/o FE & \underline{66.77\%} & \underline{69.70\%} & \underline{66.97\%} & \underline{79.68\%} & \underline{80.73\%} & \underline{80.23\%} & \underline{82.03\%} & \underline{82.23\%} & \underline{83.49\%} \\
    \mname & \textbf{66.47\%} & \textbf{68.94\%} & \textbf{66.11\%} & \textbf{79.60\%} & \textbf{80.58\%} & \textbf{80.21\%} & \textbf{81.96\%} & \textbf{81.65\%} & \textbf{83.21\%} \\
     \bottomrule
    \end{tabular}}
\end{table*}
\begin{table*}[!ht]
    \centering
    \caption{Attack Efficacy of Various Algorithms at \(\alpha=0.3, \beta=0.7\)}
    \label{tab:n7f3}
    \resizebox{0.85\linewidth}{!}{\begin{tabular}{l|ccc|ccc|ccc}
    \toprule
    Dataset  & \multicolumn{3}{c}{CITESEER} & \multicolumn{3}{|c|}{CORA} & \multicolumn{3}{c}{CORA-ML} \\ \midrule
    Target Model & \multicolumn{1}{c}{GCN} & \multicolumn{1}{c}{GAT} & \multicolumn{1}{c}{SAGE} & \multicolumn{1}{|c}{GCN} & \multicolumn{1}{c}{GAT} & \multicolumn{1}{c}{SAGE} & \multicolumn{1}{|c}{GCN} & \multicolumn{1}{c}{GAT} & \multicolumn{1}{c}{SAGE} \\ \midrule
    Clean & 70.76\% & 73.33\% & 69.99\% & 83.39\% & 84.06\% & 82.74\% & 86.90\% & 86.54\% & 86.13\% \\ \midrule
    DICE & 68.69\% & 71.67\% & 68.34\% & 82.39\% & 83.54\% & 81.53\% & 85.18\% & 85.08\% & 84.70\% \\
    EpoAtk & 71.41\% & 72.94\% & 69.10\% & 82.99\% & 83.14\% & 81.93\% & 85.41\% & 85.45\% & 85.61\% \\
    GraD & 68.73\% & 71.10\% & 67.75\% & 82.00\% & 82.44\% & 81.98\% & 84.30\% & 84.29\% & 85.16\% \\
    PGD & 70.30\% & 71.96\% & 68.05\% & 82.56\% & 82.81\% & 81.62\% & 85.33\% & 84.91\% & 84.60\% \\
    Meta-Self & 67.94\% & 71.20\% & 67.71\% & 81.82\% & 82.34\% & 81.47\% & 84.13\% & 84.44\% & 84.55\% \\
    Meta-Train & 69.99\% & 71.65\% & 68.79\% & 81.92\% & 83.25\% & 81.51\% & 84.46\% & 84.54\% & 85.18\% \\
    A-Meta-Train & 71.07\% & 72.19\% & 69.16\% & 81.84\% & 83.48\% & 81.59\% & 85.34\% & 85.20\% & 84.85\% \\
    A-Meta-Self & 68.02\% & 70.68\% & 67.92\% & 81.02\% & 81.31\% & 80.75\% & 84.04\% & 83.38\% & 84.49\% \\
    A-Meta-Both & 70.07\% & 72.65\% & 68.79\% & 82.31\% & 83.75\% & 81.51\% & 84.98\% & 85.00\% & 85.11\% \\ \midrule
    KNN-10 & 68.44\% & 70.45\% & 67.73\% & 80.64\% & \underline{81.06\%} & 81.00\% & 84.20\% & 83.64\% & 84.83\% \\
    KNN-100 & 69.16\% & 70.70\% & 68.82\% & 80.82\% & 81.44\% & 81.04\% & 83.92\% & 83.19\% & 84.26\% \\
    MEAN & 68.53\% & 70.43\% & 67.76\% & 80.61\% & 81.46\% & \underline{80.65\%} & 83.98\% & 83.42\% & 84.59\% \\ \midrule
    \mname w/o FE & \underline{67.57\%} & \underline{69.87\%} & \underline{67.11\%} & \underline{80.40\%} & 81.39\% & 80.95\% & \underline{82.87\%} & \underline{82.57\%} & \underline{83.32\%} \\
    \mname & \textbf{67.44\%} & \textbf{69.65\%} & \textbf{66.93\%} & \textbf{80.16\%} & \textbf{80.09\%} & \textbf{80.31\%} & \textbf{82.48\%} & \textbf{82.21\%} & \textbf{83.27\%} \\
     \bottomrule
    \end{tabular}}
\end{table*}

\subsection{Attack Effect Experiments}

\Cref{tab:n7f1,tab:n7f3,tab:n9f1,tab:n9f3} present the results of the attack effect experiments, where ``Clean'' indicates the accuracy of the target model without an attack.
For each attack model, lower accuracy indicates better attack performance. 
The best result in each set of experiments is highlighted in bold, while the second-best result is underlined.
In these tables, \(\alpha\) denotes the proportion of missing attributes per vertex, and \(\beta\) is the proportion of vertices with incomplete attributes.
 

\mypara{Analysis of attack performance for $\beta=0.7$.}
\Cref{tab:n7f1,tab:n7f3} present the experimental results for $\alpha = 0.1$ and $0.3$ when $\beta=0.7$.
Overall, \mname achieves the best performance. 
Relative to the model's original performance, \mname improves attack effectiveness by reducing the target model's accuracy by an average of $4.89\%$ and $4.33\%$ at $\alpha = 0.1$ and $0.3$, respectively. 
These results demonstrate that the \mname framework significantly mitigates the effects caused by incomplete information.
Compared with other existing attack methods, \mname achieves a relative improvement in average attack effectiveness of $1.53\%$ and $1.24\%$ at $\alpha=0.1$ and $0.3$, respectively. 
These results show that, compared to existing state-of-the-art methods, \mname achieves stronger attack performance on incomplete graphs.
Further analysis of the results reveals that baselines using data imputation strategies perform better than the original attack methods.
However, they still underperform when compared to \mname. 
This result suggests that conventional methods for poisoning attacks still have room for improvement, especially with incomplete graphs.
Moreover, the \mname w/o FE variant outperforms the baselines, yet \mname itself still demonstrates superior performance.
This result highlights that the Holistic Adversarial Attack module significantly enhances performance on incomplete graphs.

\mypara{Analysis of attack performance for $\beta=0.9$.}
The experimental results for $\beta=0.9$ (with more vertices having missing attributes) are shown in~\Cref{tab:n9f1,tab:n9f3}, and the overall patterns are similar to those observed for $\beta=0.7$.
When comparing the two cases, the attack performance of the baselines declines at $\beta=0.9$ due to the increased number of vertices with missing attributes. 
In contrast, with its robust attack strategy, \mname maintains strong performance.
Specifically, \mname degrades the target model's performance by $3.82\%$ and $3.88\%$ when the majority of features from $10\%$ and $30\%$ of the vertices are inaccessible.
In experiments where \mname performs better, it shows an average relative improvement of $1.03\%$ and $1.45\%$ compared to the second-best baseline at \(\alpha=0.1\) and \(\alpha=0.3\), respectively.
The experimental results for baselines using data imputation strategies are consistent with the previous analysis, showing similar performance trends.
Additionally, the \mname w/o FE variant continues to outperform the baselines but remains inferior to \mname.
This demonstrates that the feature optimization strategy employed in \mname remains effective, even when nearly all vertices experience attribute loss.

\begin{table*}[!ht]
    \centering
    \caption{Attack Efficacy of Various Algorithms at \(\alpha=0.1, \beta=0.9\)}
    \label{tab:n9f1}
    \resizebox{0.85\linewidth}{!}{
    \begin{tabular}{l|ccc|ccc|ccc}
    \toprule
    Dataset  & \multicolumn{3}{c|}{CITESEER} & \multicolumn{3}{c|}{CORA} & \multicolumn{3}{c}{CORA-ML} \\ \midrule
    Target Model & \multicolumn{1}{c}{GCN} & \multicolumn{1}{c}{GAT} & \multicolumn{1}{c|}{SAGE} & \multicolumn{1}{c}{GCN} & \multicolumn{1}{c}{GAT} & \multicolumn{1}{c|}{SAGE} & \multicolumn{1}{c}{GCN} & \multicolumn{1}{c}{GAT} & \multicolumn{1}{c}{SAGE} \\ \midrule
    Clean & 70.76\% & 73.33\% & 69.99\% & 83.39\% & 84.06\% & 82.74\% & 86.90\% & 86.54\% & 86.13\% \\ \midrule
    DICE & 70.32\% & 71.14\% & \underline{68.02\%} & 82.05\% & 82.48\% & 81.73\% & 85.19\% & 84.44\% & 84.83\% \\
    EpoAtk & 70.58\% & 72.02\% & 68.67\% & 81.47\% & 82.64\% & 81.53\% & 85.79\% & 85.24\% & 85.53\% \\
    GraD & 69.63\% & 71.88\% & 68.64\% & 81.76\% & 82.96\% & 82.18\% & 82.90\% & 83.41\% & 84.77\% \\
    PGD & 71.60\% & 72.59\% & 69.28\% & 82.61\% & 83.38\% & 81.83\% & 85.90\% & 85.41\% & 84.90\% \\
    Meta-Self & 69.07\% & 71.79\% & 68.65\% & 81.07\% & 82.60\% & 81.55\% & 83.12\% & 83.44\% & 84.34\% \\
    Meta-Train & 70.35\% & 71.88\% & 68.67\% & 82.05\% & 82.81\% & 81.84\% & 84.25\% & 84.26\% & 84.84\% \\
    A-Meta-Train & 71.33\% & 73.21\% & 70.25\% & 82.94\% & 83.65\% & 82.51\% & 86.03\% & 85.44\% & 85.41\% \\
    A-Meta-Self & 70.67\% & 72.19\% & 69.91\% & 81.56\% & 82.23\% & 81.16\% & 85.56\% & 84.38\% & 84.65\% \\
    A-Meta-Both & 72.02\% & 73.07\% & 68.92\% & 82.77\% & 83.63\% & 81.70\% & 86.14\% & 85.43\% & 85.57\% \\ \midrule
    KNN-10 & 70.55\% & 72.88\% & 69.68\% & 82.12\% & 82.48\% & 81.40\% & 85.65\% & 84.76\% & 85.41\% \\
    KNN-100 & 70.25\% & 72.88\% & 69.56\% & 81.34\% & \underline{81.35\%} & 81.45\% & 85.59\% & 84.50\% & 84.95\% \\
    MEAN & 70.05\% & 72.34\% & 68.74\% & 81.21\% & 81.87\% & 81.03\% & 85.55\% & 84.81\% & 85.10\% \\ \midrule
    \mname w/o FE & \underline{68.62\%} & \underline{70.07\%} & 68.16\% & \underline{80.91\%} & 81.64\% & \textbf{80.67\%} & 83.31\% & \textbf{82.22\%} & \underline{83.87\%} \\
    \mname & \textbf{67.18\%} & \textbf{70.02\%} & \textbf{67.09\%} & \textbf{80.85\%} & \textbf{81.31\%} & \underline{80.91\%} & \textbf{82.82\%} & \underline{82.52\%} & \textbf{83.67\%} \\
    \bottomrule
    \end{tabular}}
\end{table*}
\begin{table*}[!ht]
    \centering
    \caption{Attack Efficacy of Various Algorithms at \(\alpha=0.3, \beta=0.9\)}
    \label{tab:n9f3}
    \resizebox{0.85\linewidth}{!}{
        \begin{tabular}{l|ccc|ccc|ccc}
            \toprule
            Dataset & \multicolumn{3}{c|}{CITESEER} & \multicolumn{3}{c|}{CORA} & \multicolumn{3}{c}{CORA-ML} \\ 
            \midrule
            Target Model & \multicolumn{1}{c}{GCN} & \multicolumn{1}{c}{GAT} & \multicolumn{1}{c|}{SAGE} & \multicolumn{1}{c}{GCN} & \multicolumn{1}{c}{GAT} & \multicolumn{1}{c|}{SAGE} & \multicolumn{1}{c}{GCN} & \multicolumn{1}{c}{GAT} & \multicolumn{1}{c}{SAGE} \\ 
            \midrule
            Clean & 70.76\% & 73.33\% & 69.99\% & 83.39\% & 84.06\% & 82.74\% & 86.90\% & 86.54\% & 86.13\% \\ 
            \midrule
            DICE & 69.91\% & 71.99\% & 68.96\% & 82.07\% & 82.62\% & 81.77\% & 85.35\% & 85.15\% & 85.13\% \\
            EpoAtk & 70.17\% & 71.99\% & 67.37\% & 83.41\% & 83.40\% & 82.56\% & 85.02\% & 85.44\% & 84.95\% \\
            GraD & 69.15\% & 72.33\% & 69.81\% & 81.84\% & 82.88\% & 81.89\% & 83.86\% & 83.79\% & 84.24\% \\
            PGD & 69.30\% & 71.70\% & 68.86\% & 82.69\% & 83.22\% & 81.68\% & 85.85\% & 85.31\% & 84.68\% \\
            Meta-Self & 70.19\% & 70.99\% & 69.02\% & 81.90\% & 82.99\% & 81.95\% & \textbf{83.59\%} & 83.84\% & 84.81\% \\
            Meta-Train & 71.15\% & 73.26\% & 69.65\% & 82.56\% & 83.28\% & 82.51\% & 85.82\% & 85.73\% & 84.99\% \\
            A-Meta-Train & 71.24\% & 73.12\% & 69.39\% & 82.55\% & 83.99\% & 82.07\% & 85.99\% & 85.41\% & 85.24\% \\
            A-Meta-Self & 70.51\% & 72.57\% & 69.23\% & 82.55\% & 83.13\% & 81.39\% & 85.69\% & 84.47\% & 84.43\% \\
            A-Meta-Both & 71.42\% & 72.14\% & 69.44\% & 82.92\% & 83.76\% & 81.80\% & 85.87\% & 85.34\% & 85.39\% \\
            \midrule
            KNN-10 & 71.23\% & 72.81\% & 68.54\% & 82.39\% & 82.56\% & 81.94\% & 85.78\% & 84.84\% & 84.96\% \\
            KNN-100 & 71.29\% & 72.93\% & 69.96\% & 82.33\% & 83.02\% & 81.55\% & 85.45\% & 84.92\% & 85.16\% \\
            MEAN & 71.27\% & 73.10\% & 70.56\% & 82.61\% & 82.56\% & 81.41\% & 85.87\% & 84.51\% & 85.00\% \\
            \midrule
            \mname w/o FE & \underline{67.66\%} & \underline{70.93\%} & \underline{67.22\%} & \underline{80.43\%} & \underline{81.96\%} & \underline{80.53\%} & 83.67\% & \textbf{82.45\%} & \underline{83.91\%} \\
            \mname & \textbf{67.55\%} & \textbf{70.24\%} & \textbf{67.04\%} & \textbf{80.17\%} & \textbf{81.10\%} & \textbf{80.31\%} & \underline{83.61\%} & \underline{82.50\%} & \textbf{83.38\%} \\
            \bottomrule
        \end{tabular}
    }
\end{table*}
By comparing the results across these experimental results, it is evident that existing methods exhibit instability in scenarios involving missing attributes. 
The state-of-the-art method A-Meta-Self maintains some effectiveness at \( \beta = 0.7 \).
However, its performance significantly deteriorates when \( \beta = 0.9 \). 
The reason is that the self-learning component of the model becomes ineffective when a substantial number of vertices lack attributes, leading to a considerable drop in attack performance.

\subsection{Ablation Study}
\label{sec:abl}
We conducted ablation study experiments to evaluate the model's ability to utilize information.
The results are shown in~\Cref{tab:ab}, using error rate as the evaluation metric.
The error rate is defined as the proportion of incorrect predictions or misclassifications, calculated as the number of errors divided by the total number of predictions. 
A lower error rate indicates stronger model capability.
We selected $\alpha=0.1$ and $\beta=0.7$ to test the module's effectiveness in a relaxed environment, appropriately reducing $K$ in the ablation models to ensure effective attacks.
We began by examining the self-learning component in the state-of-the-art A-Meta-Self baseline, referred to as AMS-SL. 
Subsequently, we conducted individual ablation experiments on the Bifocal Feature Processor (BFP) and the local and global attention components within the Local-global Aggregation module.

\begin{table*}[!tb]
    \centering
    \caption{Ablation Study Results}
    \label{tab:ab}
    \begin{tabular}{ccc|ccc}
    \toprule
          &       &       & CITESEER & CORA & CORA-ML \\ \midrule
    \multicolumn{3}{c|}{AMS-SL} & 50.41\% & 47.49\% & 56.89\% \\ \midrule
    Global & Local & BFP  &       &       &  \\ \midrule
          &       &       & 36.32\% & 23.09\% & 29.43\% \\
    \checkmark     &       &       & 33.07\% & 18.70\% & 23.56\% \\
          & \checkmark     &       & 32.47\% & 18.48\% & 23.56\% \\
    \checkmark     & \checkmark     &       & 31.22\% & 17.46\% & 19.97\% \\
    \checkmark     &       & \checkmark     & 30.59\% & 16.24\% & 20.27\% \\
          & \checkmark     & \checkmark     & 30.31\% & 16.07\% & 19.69\% \\
    \checkmark     & \checkmark     & \checkmark     & \textbf{28.88\%} & \textbf{15.79\%} & \textbf{15.67\%} \\ \bottomrule
    \end{tabular}
\end{table*}

The experimental results show that the complete model performs best across all three datasets. 
More specifically, AMS-SL exhibits unsatisfactory accuracy across all datasets because it is designed for complete graphs and has difficulty accurately capturing the correlations between incomplete attributes and the graph structure.
Further analysis of the experimental results reveals that the inclusion of any component enhances the performance of the Depth-plus GNN module, highlighting the effectiveness of both BFP and Local-global Aggregation module.
Moreover, utilizing the BFP component leads to performance improvements in both global and local attention, as BFP enhances the noise-reduction capabilities of these attention mechanisms. 
Additionally, the performance improvement from the local attention component surpasses that of the global attention component. 
This is because local attention focuses on preserving crucial features during the propagation process, which plays a key role in enhancing model performance.

\subsection{Further Analysis}
\label{sec:furtheranalyses}
We conducted three additional sets of experiments on the CORA dataset to further assess the performance of \mname.

\begin{figure}[!tb]
  \centering
  \includegraphics[width=0.6\linewidth]{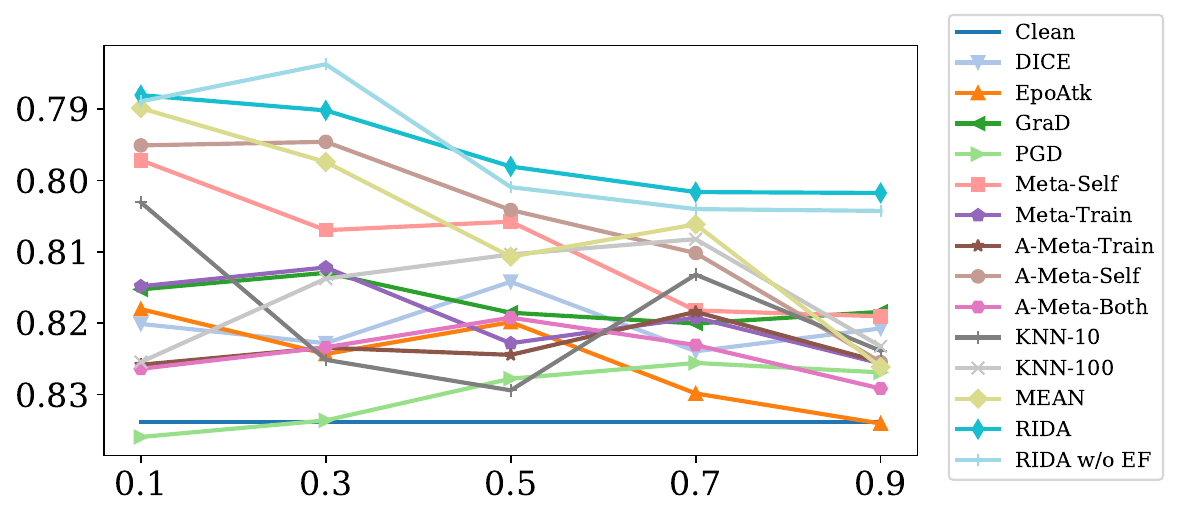}
  \caption{Attack Performance of Models Across Varying \(\beta\)}
  \label{fig:nrates}
\end{figure}

\mypara{Robustness analysis.}
In this set of experiments, we select GCN as the target model.
The attack performance of all methods was evaluated across various values of \( \beta \) when \( \alpha = 0.3 \) to assess their robustness.
The metric used to assess the performance of the attacked model is accuracy, where a lower value indicates better attack performance. 
The results are shown in~\Cref{fig:nrates}, with the y-axis inverted for clarity. 
Higher positions indicate better attack performance.
The results show that both \mname and \mname w/o EF achieve the best attack performance across all scenarios. 
Notably, \mname maintains outstanding attack performance even when a significant number of vertices lack attributes. 
Compared to \mname w/o EF, \mname demonstrates superior attack performance in almost all scenarios. 
The experiment confirms the effectiveness of the optimization strategy for \(\bm{X}_s\), revealing that guiding the model to focus on information across the entire graph during the attack process further improves its robustness.
In conclusion, these results highlight the capabilities of \mname in attacking incomplete graphs.

\begin{figure}[!tb]
  \centering
  \subfigure[Performance on varying $\alpha$, $\beta$: y-axis is accuracy, x-axis is proportion distance.]{
\includegraphics[width=0.3\linewidth]{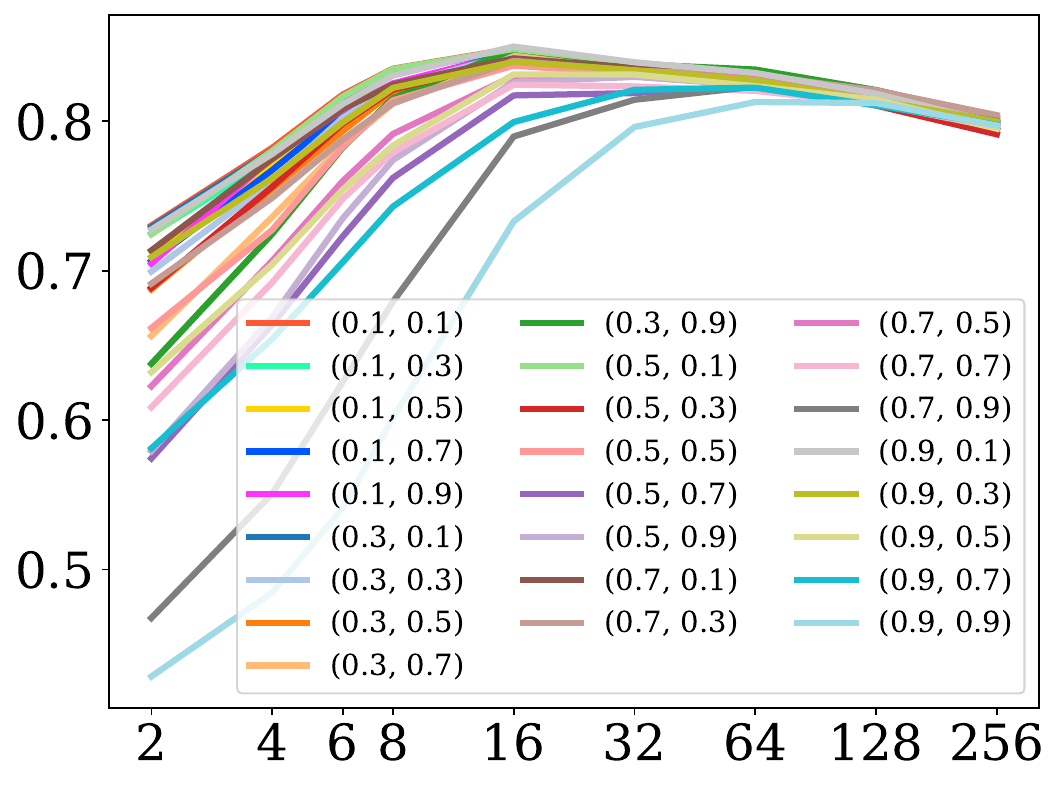}
   \label{fig:layer}
  }
  \subfigure[Propagation when $K=16$: color indicates the weight between vertices.]{
\includegraphics[width=0.3\linewidth]{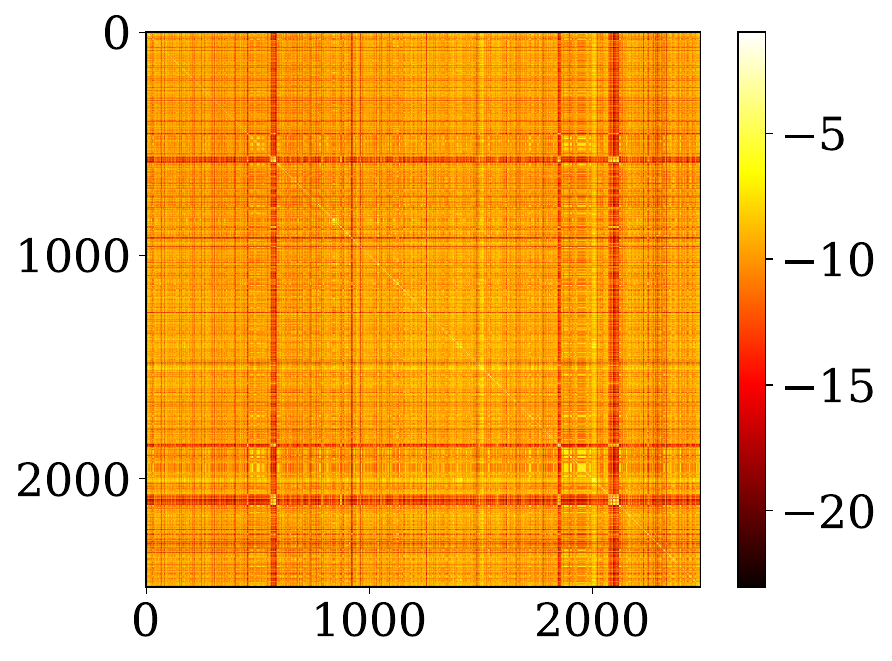}
   \label{fig:heat}
  }
  \caption{In-depth Analysis}
  
\end{figure}

\mypara{Analysis of surrogate model optimization.}
In this set of experiments, we further analyzed how increasing the propagation distance helps optimize the surrogate model in \mname.
\Cref{fig:layer} illustrates \mname's ability to utilize information under scenarios with \(\alpha=30\%\) and \(\beta \in [10\%, 90\%]\). 
The y-axis represents model effectiveness, measured by accuracy metric, while the x-axis represents the propagation distance \(K\). 
Higher values indicate better performance.
The results demonstrate that increasing the propagation distance \(K\) within a certain range can significantly improve performance across all scenarios.
As the extent of missing attributes increases, a larger \(K\) is required for optimal performance.
These experiments highlight the importance of leveraging attributes from more distant vertices on attribute-incomplete graphs.
Notably, with the help of the decay and attention components, the model maintains stable performance as \(K\) increases. 
This is because these components enable the model to discern the importance of features, ensuring valid propagation of distant information and improving performance in scenarios with missing attributes.


\mypara{Analysis of attribute propagation process.}
To conduct an in-depth analysis of the effectiveness of the model's attribute propagation behavior, we examined the performance of the process at \(K=16\), which allows the model to propagate information across a relatively large number of hops. 
After applying a logarithmic transformation, the model's final propagation matrix is visualized as a heatmap in \Cref{fig:heat}, offering a clear representation of how information flows through the graph.
The heatmap reveals that information is transmitted between nearly every pair of vertices in the graph, indicating an extensive exchange of data.
This widespread propagation occurs because the distance model collects information beyond the graph's diameter.
By transmitting comprehensive data across the entire graph, \mname becomes less susceptible to the impact of missing attributes, thereby ensuring that it can still perform effective attacks on incomplete graphs.
\section{Related Works}
Poisoning attacks are a specific type of adversarial attack.
Adversarial attacks in graph neural networks (GNNs) are complex and diverse, and they can be categorized by several factors: the task level (e.g., vertex-level~\citep{zugner2018adversarial,wu2019adversarial,chang2020restricted,entezari2020all,liu2023everything} or graph-level~\citep{dai2018adversarial,tang2020adversarial}), the purpose of the attack (targeted~\citep{zugner2018adversarial,xu2019topology} or untargeted~\citep{zügner2018adversarial}), the stage of the attack (poisoning~\citep{liu2019unified,sun2020adversarial,dai2023unnoticeable} or evasion~\citep{chang2020restricted,wang2020evasion}), the attacker’s knowledge of the model (white-box~\citep{wang2019attacking,wu2019adversarial}, black-box~\citep{dai2018adversarial,ma2020towards,ma2022adversarial}, or gray-box~\citep{zügner2018adversarial,bojchevski2019adversarial,sun2020adversarial}), and the nature of the perturbation (e.g., modifying vertex features~\citep{ma2020towards,zugner2018adversarial}, altering the graph’s structure~\citep{liu2022gradients,zügner2018adversarial}, or adding new vertices~\citep{sun2020adversarial,tao2021adversarial,zou2021tdgia}).
Compared to modifying vertex features or introducing new vertex, perturbing edges is easier to execute and requires less effort for the attacker. 
Since real-world graph structures are often sparse, altering a small number of edges can significantly mislead GNN models' perception of critical substructures, making global attacks more feasible.

Previous studies on vertex-level untargeted gray-box poisoning attacks on graph structures have introduced effective methods~\citep{jin2021adversarial,liu2022gradients,liu2022towards,wu2019adversarial,xu2019topology,zügner2018adversarial}.
For instance, meta-based attacks~\citep{zügner2018adversarial} leverage meta-learning techniques by treating the input data as a hyperparameter, enabling the model to adaptively optimize for attack objectives. 
The PGD method~\citep{xu2019topology} employs a powerful first-order optimization technique, which efficiently identifies perturbation strategies.
Moreover, GraD~\citep{liu2022towards} introduces a novel objective function to mitigate gradient bias, leading to more precise attack outcomes.
Despite their strengths, these methods are designed under the ideal assumption that attackers have access to complete graph information. 
As a result, they lack specific optimizations for scenarios with missing attributes, limiting their ability to effectively identify the most impactful perturbation targets in realistic attack scenarios.
Although some existing studies focus on executing classic graph tasks on incomplete graphs~\citep{zhang2023amgcl,lin2022dual,zheng2023finding,liu2023learning}, these approaches are not designed for attack scenarios. 
To the best of our knowledge, \mname is the first method to achieve robust gray-box poisoning attacks on incomplete graphs.

\section{Conclusion}
This paper introduces \mname, an innovative model designed for gray-box poisoning attacks for GNNs on incomplete attributed graphs. 
\mname addresses the challenges of such attack scenarios by enhancing the model's ability to propagate and aggregate information from distant vertices while providing optimized strategies for the attack process.
The model consists of three main modules: the Depth-plus GNN module, which ensures comprehensive information propagation; the Local-global Aggregation module, which refines the aggregation process through attention mechanisms; and the Holistic Adversarial Attack module, which leverages long-range propagation traces to optimize and execute high-performance attacks.
Extensive experiments on three real-world datasets, compared against nine state-of-the-art baselines, demonstrate that \mname excels in scenarios with missing attributes, validating its effectiveness and offering valuable insights for improving the security and robustness of GNNs in practical applications.

\printcredits

\bibliographystyle{apacite}
\bibliography{citations}

\end{document}